\pdfoutput=1
\documentclass{article}
\usepackage[final]{colm2026_conference}

\usepackage{microtype}
\usepackage{hyperref}
\usepackage{url}
\usepackage{booktabs}

\usepackage[T1]{fontenc}

\usepackage{amsmath,amssymb,mathtools,amsthm}

\usepackage{graphicx}
\usepackage{subcaption}
\usepackage{multirow}
\usepackage{array}
\usepackage{tabularx}
\usepackage{longtable}
\usepackage{adjustbox}
\usepackage{wrapfig}

\usepackage{enumitem}
\usepackage[table]{xcolor}
\usepackage[most]{tcolorbox}

\usepackage{titletoc}

\usepackage[capitalize,noabbrev]{cleveref}

\usepackage{pifont}

\definecolor{ICMLBlue}{rgb}{0.0, 0.0, 0.5}
\definecolor{CiteTeal}{HTML}{008080}
\definecolor{lightcrimson}{rgb}{0.93, 0.16, 0.51}
\definecolor{LinkRed}{rgb}{0.768, 0.054, 0.054}

\hypersetup{
    colorlinks=true,
    linkcolor=ICMLBlue,      
    citecolor=CiteTeal,      
    urlcolor=ICMLBlue,   
    linktoc=page
}

\definecolor{lightgray}{gray}{0.9}
\definecolor{lightblue}{RGB}{230,245,255}

\theoremstyle{plain}

\theoremstyle{definition}

\theoremstyle{remark}

\titlecontents{section}[1.5em]
    {\addvspace{6pt}\bfseries}
    {\contentslabel{1.5em}}
    {\hspace*{-1.5em}}
    {\titlerule*[0.5pc]{.}\contentspage}

\titlecontents{subsection}[3.8em]
    {}
    {\contentslabel{2.3em}}
    {\hspace*{-2.3em}}
    {\titlerule*[0.5pc]{.}\contentspage}

\makeatletter
\@ifpackageloaded{algorithm}{
    
}{}
\makeatother

\tcbuselibrary{breakable,skins,listings}

\definecolor{BoxGrayBack}{HTML}{F8F8F8}
\definecolor{BoxGrayFrame}{HTML}{B8B8B8}
\definecolor{BoxBlueBack}{HTML}{F5F9FF}
\definecolor{BoxBlueFrame}{HTML}{2D5DA8}
\definecolor{BoxGreenBack}{HTML}{F6FFF8}
\definecolor{BoxGreenFrame}{HTML}{2E7D32}

\newtcolorbox{detailbox}[2][]{
  enhanced,
  breakable,
  colback=BoxGrayBack,
  colframe=BoxGrayFrame,
  boxrule=0.5pt,
  arc=2pt,
  left=6pt,
  right=6pt,
  top=5pt,
  bottom=5pt,
  fonttitle=\bfseries,
  coltitle=black,
  title={#2},
  #1
}

\newtcblisting{promptbox}[2][]{
  enhanced,
  breakable,
  listing only,
  colback=BoxBlueBack,
  colframe=BoxBlueFrame,
  boxrule=0.6pt,
  arc=2pt,
  left=6pt,
  right=6pt,
  top=5pt,
  bottom=5pt,
  fonttitle=\bfseries,
  coltitle=white,
  colbacktitle=BoxBlueFrame,
  title={#2},
  listing options={
    basicstyle=\ttfamily\footnotesize,
    breaklines=true,
    breakatwhitespace=false,
    columns=fullflexible,
    keepspaces=true,
    showstringspaces=false,
    upquote=true
  },
  #1
}

\newtcolorbox{implementationbox}[2][]{
  enhanced,
  breakable,
  colback=BoxGreenBack,
  colframe=BoxGreenFrame,
  boxrule=0.6pt,
  arc=2pt,
  left=6pt,
  right=6pt,
  top=5pt,
  bottom=5pt,
  fonttitle=\bfseries,
  coltitle=white,
  colbacktitle=BoxGreenFrame,
  title={#2},
  #1
}

\newtcolorbox{takeawaybox}{
  colback=red!4,
  colframe=red!60!black,
  boxrule=0.65pt,
  arc=1.5pt,
  left=6pt,
  right=6pt,
  top=5pt,
  bottom=5pt
}

\hypersetup{
    pdftitle={Towards Reliable Context Compression for Long-Horizon Agents: An Empirical Study of Execution Instability},
    pdfauthor={Guanghui Min, Liang Wu, Mayank Darbari, Chen Chen, Liangjie Hong},
    pdfkeywords={LLM agents, context compression, long-horizon, prompt optimization, AppWorld},
}

\title{
Towards Reliable Context Compression for Long-Horizon Agents:
An Empirical Study of Execution Instability
}

\author{
Guanghui Min$^{1,2}$,
Liang Wu$^{2}$,
Mayank Darbari$^{2}$,
Chen Chen$^{1}$,
Liangjie Hong$^{2}$
\\[0.5em]
$^{1}$Department of Computer Science, University of Virginia,
Charlottesville, VA, USA
\\
$^{2}$Nokia, Sunnyvale, CA, USA
\\[0.5em]
\texttt{\{jjm8vr, zrh6du\}@virginia.edu}
\\
\texttt{\{liang.wu, mayank.darbari, liangjie.hong\}@nokia.com}
}

\begin{document}

\maketitle

\begin{abstract}
Recurrent context compression controls context growth in long-horizon agents, but its behavioral effects remain poorly understood. In this preliminary empirical study, we show that compression can weaken the influence of recent interactions, increasing blocked actions, repeated exploration, and instability across runs. Motivated by these observations, we introduce \texttt{TRACE}, a verifier-guided framework that evaluates individual compaction events through paired closed-loop continuations from the same environment state and uses summary preferences to optimize a natural-language compression prompt while keeping all models frozen. Initial results on AppWorld show improvements over existing compression baselines in task performance, multi-run reliability, and context--execution efficiency. These findings provide early evidence for boundary-local evaluation as a promising direction for reliable agent context compression. Our code is in \url{https://github.com/nokia-applied-research/Trace}.
\end{abstract}

\section{Introduction}
Large language models increasingly operate over long, evolving contexts rather than isolated prompts. Such settings include interactive agents that interleave reasoning, actions, observations, and plan revision~\citep{yao2023react,shinn2023reflexion,wang2024codeact}; browser, application, API, and office environments that require extended interaction with external systems~\citep{zhou2024webarena,trivedi2024appworld,wang2024officebench}; policy-constrained conversational workflows; and retrieval-intensive research tasks that accumulate evidence over multiple steps. In all of these settings, the context grows with every tool output, retrieved source, intermediate decision, user clarification, and partial result. Repeatedly supplying the full history increases inference cost and peak-context pressure, while making the information needed for the next decision increasingly difficult to locate. Context compression is therefore essential for scalable long-horizon LLM systems.

Prior work has progressively moved from document-level prompt compression to trajectory-aware context management. General prompt-compression methods prune, rewrite, or distill input tokens to reduce inference cost while preserving semantic content or downstream task quality~\citep{li2023selective,jiang2023llmlingua,jiang2024longllmlingua,pan2024llmlingua,xu2024recomp,shandilya2025taco}. For long-horizon agents, recent methods recognize that context contains more than ordinary prose: it records action--observation histories, intermediate plans, and evolving interaction state. ReSum and SUPO therefore co-optimize summarization with downstream agent behavior~\citep{wu2025resum,lu2025supo}; ACON improves compression guidelines by contrasting successful full-context trajectories with failed compressed ones~\citep{kang2025acon}; and practical systems compact long sessions into structured natural-language checkpoints~\citep{openclaw_compaction,openclaw_pruning,hermes_context_compression}. These advances establish that agent context should be treated differently from a static document.

Nevertheless, these approaches share an implicit substitution assumption: \textit{once a shorter checkpoint retains the task-relevant content, it can stand in for the original interaction history}. This assumption overlooks the directional role of history in long-horizon execution. A raw trajectory does not merely record facts; its ordered action--observation sequence anchors what has already been completed, where execution currently stands, and whether the task is ready to terminate. Summarization can flatten this directed process into a declarative account of past progress and future plans. As a result, even when salient entities and progress labels are retained, a frozen agent may lose its local position in the trajectory, replay completed actions, or continue acting beyond the terminal frontier. The central challenge is therefore to compress history while preserving a representation from which the frozen agent can reliably recognize its current execution state.

We propose \texttt{TRACE}
(\textbf{\underline{T}}rajectory-\textbf{\underline{R}}elative
\textbf{\underline{A}}gent
\textbf{\underline{C}}ontext Compr\textbf{\underline{E}}ssion), a
verifier-guided framework for optimizing recurrent context compression for
frozen long-horizon agents. Rather than inferring compression quality from
terminal task outcomes, \texttt{TRACE} evaluates each compaction boundary
directly. From the same environment state, it compares paired closed-loop
continuations before and after compression and measures the additional blocked
or repeated exploration induced by the summary. These boundary-local scores
produce preferences between candidate summaries. A frozen proposer observes
only which summary is preferred---not the subsequent actions, observations, or
error signals---and uses these preferences to revise the natural-language
compression template. The compressor model, downstream agent, tools, and
decoding configuration remain fixed throughout.

Our contributions are threefold:
\begin{itemize}[leftmargin=*]
\item \textbf{Behavioral diagnosis.}
We show that recurrent compression can attenuate the effect of recent
interactions, increasing blocked execution and repeated exploration while
reducing multi-run reliability.
\item \textbf{Boundary-local compression optimization.}
We introduce a paired closed-loop verifier that separates
compression-induced execution regressions from the frozen agent's intrinsic
behavioral variability, and use its summary preferences to optimize the
compression template without exposing rollout details to the proposer.
\item \textbf{Empirical evaluation.}
On AppWorld, \texttt{TRACE} consistently outperforms existing compression
baselines across task difficulty levels, improves repeated-run reliability, and
keeps average execution steps close to full context while substantially reducing
peak context usage. Moreover, the template optimized with MiniMax-M3 transfers
to Kimi-K2.7-Code without further optimization, outperforming all compressed
baselines and even exceeding full context in overall accuracy and Pass$^2$.
\end{itemize}

\section{Preliminaries}\label{sec:prelim}
\subsection{Context Compression for Frozen Long-Horizon Agents}
\label{sec:prelim-setup}

\noindent\textbf{Notations.}
We study context compression for a fixed downstream agent $\mathcal{M}$ interacting with an external environment $\mathcal{E}$.
The agent is fixed in the sense that its language model, system prompt, action interface, output parser, and decoding procedure are not modified.
Let $\mathcal{D}=\{u_i\}_{i=1}^{|\mathcal{D}|}$ denote a distribution of long-horizon tasks, where each task begins with an instruction $u$.
A rollout is denoted by $\tau=(u,a_1,o_1,\ldots,a_T,o_T)$.
Before decision step $t$, the complete interaction history is $h_t=(u,a_1,o_1,\ldots,a_{t-1},o_{t-1})$, where $a_i$ is an agent action and $o_i=\mathcal{E}(a_i)$ is the corresponding environment observation.
We distinguish the complete interaction history $h_t$ from the \textit{agent-visible context} $z_t$ supplied to the frozen agent.
Given $z_t$, the agent generates $a_t\sim\mathcal{M}(\cdot\mid z_t)$ and receives $o_t=\mathcal{E}(a_t)$.
Full-context execution uses $z_t=h_t$.
Since $|h_t|$ generally grows with $t$, the complete history can eventually exceed the practical context budget of long-horizon execution.

\subsection{Problem Definition}
\label{subsec:problem_def}

Long-horizon agents must periodically replace their growing working context with a bounded textual representation~\citep{kang2025acon,lu2025supo,AG2_2024,hermes_context_compression,openclaw_compaction}.
Let $\mathcal{C}$ denote a compressor and $B$ denote the context budget.
Given the current agent-visible context $z_t$, action $a_t$, and observation $o_t$, define the pre-compaction context as $\bar z_{t+1}=z_t\oplus(a_t,o_t)$, where $\oplus$ denotes textual concatenation.
The context supplied at the next decision step is recursively updated as

\begin{equation}
z_{t+1}=
\begin{cases}
\bar z_{t+1}, & |\bar z_{t+1}|\leq B,\\
\mathcal{C}(\bar z_{t+1};B), & |\bar z_{t+1}|>B.
\end{cases}
\label{eq:context_transition}
\end{equation}

Thus, after a compaction event, the replacement context $z_{t+1}$ becomes the only historical record available to the frozen agent and to subsequent compaction steps.
For a compressor $\mathcal{C}$, let $p_{\mathcal{C},B}(\tau\mid u)$ denote the rollout distribution induced by the frozen agent $\mathcal{M}$, environment $\mathcal{E}$, and the recursive context transition in Equation~\ref{eq:context_transition}.
Let $R(\tau)$ denote the environment-defined terminal reward.

\section{Empirical Study: Behavioral Effects of Context Compression}\label{sec:motivation}
In this section, we first show that recurrent compression degrades task success and reliability, especially over longer trajectories. We then trace this degradation to execution-state mislocalization and examine how it manifests as blocked execution and repeated exploration after compaction.

\subsection{Repeated Compression Degrades Agent Behavior}
\label{sec:repeated-compression}

\begin{figure*}[t]
    \centering
\begin{subfigure}[t]{0.315\textwidth}
    \centering
    \includegraphics[width=\linewidth]{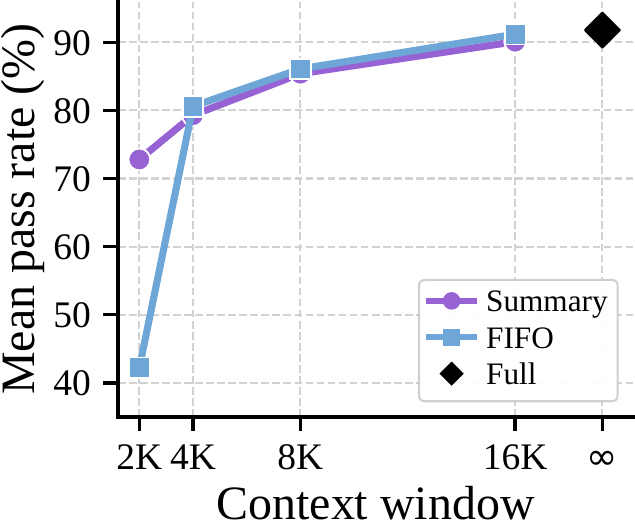}
    \caption{Compression performance.}
    \label{fig:compaction_budget}
\end{subfigure}
\hfill
\begin{subfigure}[t]{0.315\textwidth}
    \centering
    \includegraphics[width=\linewidth]{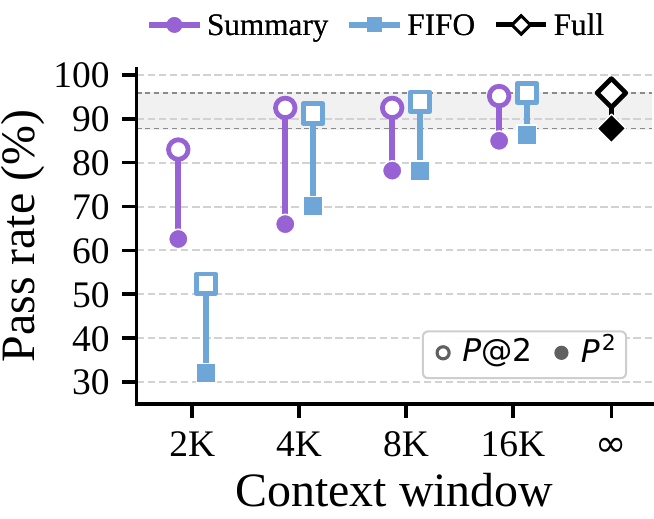}
    \caption{Cross-run behavioral instability.}
    \label{fig:compaction_stability}
\end{subfigure}
\hfill
\begin{subfigure}[t]{0.315\textwidth}
    \centering
    \includegraphics[width=\linewidth]{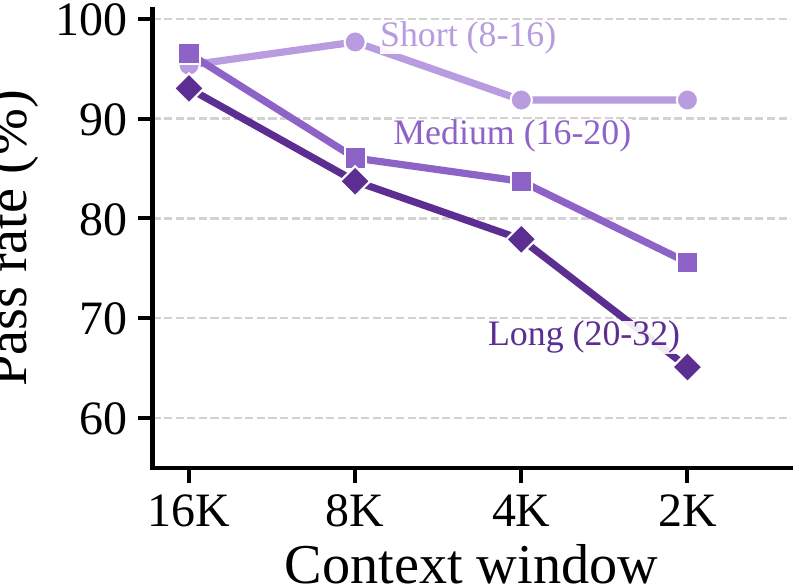}
    \caption{Long-horizon degradation.}
    \label{fig:compaction_horizon}
\end{subfigure}
    \vspace{-2mm}
\caption{\textbf{Repeated context replacement degrades agent behavior despite refetchable information.}
AppWorld allows agents to re-query persistent application state, so previously observed information remains recoverable after compaction.
(a) Mean pass rate declines as the context budget shrinks.
(b) The widening gap between $P@2$ and $P^2$ indicates reduced reliability across repeated runs.
(c) Even with summary-based compaction, degradation is substantially sharper for tasks with longer full-context reference horizons.}
    \label{fig:compaction_behavior}
    \vspace{-5mm}
\end{figure*}

Equation~\eqref{eq:context_transition} makes context compression a recurrent intervention: once the working context exceeds the budget, the replacement context becomes the input to both subsequent decisions and later compaction events. Its quality therefore cannot be assessed from a single replacement in isolation. A locally plausible summary may still alter the downstream rollout once it is repeatedly reused as the agent's working context.

We evaluate this effect on the 147-task AppWorld train--development split~\citep{trivedi2024appworld}. AppWorld is a stateful API-use benchmark: agents interact with persistent simulated applications through API calls, and earlier observations can be recovered by re-querying the underlying application state. This setting differs from knowledge-intensive tasks, where summary-based compaction is expected to outperform FIFO truncation because it preserves facts that would otherwise be permanently discarded. In AppWorld, removing an observation from the working context does not necessarily make its content permanently inaccessible. We use MiniMax-M3 as both compressor and downstream agent~\citep{minimax2026m3}. Our evaluation harness adapts OpenClaw's recurrent compaction loop\footnote{Adapted from the OpenClaw agent-core harness,
\url{https://github.com/openclaw/openclaw/blob/0e7b5c34292cc28707a0e5a0b730cff295ef0f8a/packages/agent-core/src/harness/compaction/compaction.ts}
(commit \texttt{0e7b5c34292cc28707a0e5a0b730cff295ef0f8a}).}~\citep{openclaw_compaction} . This common runtime pattern is also used in other agentic frameworks such as AG2 and Hermes Agent~\citep{AG2_2024,hermes_context_compression}.

Figure~\ref{fig:compaction_behavior} reveals a surprising budget-dependent pattern. Summary-based compaction does not uniformly outperform FIFO truncation: at $16$K, $8$K, and $4$K, FIFO matches or slightly exceeds summary-based execution. Their difference emerges only under the tightest budget, where summary-based compaction reaches $72.8\%$ while FIFO drops to $42.2\%$. Thus, in this information-refetchable setting, the advantage of summarization over recency-only truncation appears only under severe compression. Both policies nevertheless remain below full-context performance.

Mean pass rate alone does not fully characterize this degradation. Following the repeated-run reliability metrics used in $\tau$-bench and $\tau^2$-bench~\citep{yao2024tau,barres2025tau}, $P^k$ denotes the fraction of tasks solved in all $k$ independent runs, whereas $P@k$ denotes the fraction solved in at least one of the $k$ runs. We use $k=2$ throughout. Under stronger compression, the gap between $P@2$ and $P^2$ widens substantially. Compression therefore does not merely remove a subset of tasks from the agent's reach: it converts some tasks that were reliably solved into tasks that are only intermittently solved under the same runtime condition.

This instability is more pronounced on tasks with longer full-context execution horizons, measured by the median number of steps in successful full-context runs. Even under summary-based compaction, longer-horizon tasks degrade more sharply as the budget shrinks, consistent with an accumulating effect of repeated context replacement.

Together, these results suggest that successful compression depends not only on preserving globally relevant information, but also on maintaining the local continuity that anchors the agent's current execution position. FIFO preserves recent action--observation continuity despite discarding most earlier history, whereas repeated summary replacement can progressively weaken this local anchor over longer trajectories. We next isolate this effect at matched decision points.

\begin{takeawaybox}
    \textbf{Takeaway:} Compressor quality is better reflected by multi-run stability than by single-run performance alone.
\end{takeawaybox}

\subsection{Beyond Information Loss: Execution-State Mislocalization}
\label{sec:agent-mislocalization}

\begin{wrapfigure}{r}{0.5\textwidth}
    \centering
    \vspace{-6mm}
    \includegraphics[width=\linewidth]
{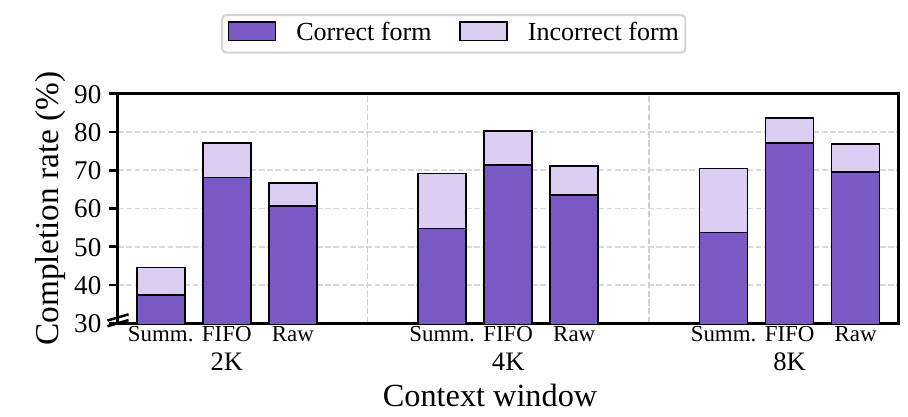}
    \vspace{-5mm}
    \caption{\textbf{Terminal completion.}
    Total height denotes the termination rate; dark segments denote
    termination in the required form.}
    \label{fig:terminal-completion}
    \vspace{-5mm}
\end{wrapfigure}

The degradation in Section~\ref{sec:repeated-compression} is often attributed
to the loss of facts, variables, or task progress during summarization. This account is incomplete in our setting: earlier
observations remain re-queryable in AppWorld, while FIFO truncation remains
competitive at moderate budgets despite discarding most earlier history.
We therefore ask whether compression also weakens the agent's ability to
recover its current execution state---what has been completed, what remains
actionable, and whether execution should continue or terminate.

\noindent\textbf{Summary Replacement Disrupts Terminal Completion.}
We first probe the final decision before termination. For each trajectory with
at least one compaction, we hold the task, environment, agent header, and
decision point fixed, and compare the actual summary context, FIFO truncation
at the same budget, and the full pre-terminal history. All turns are
reconstructed in their native interaction format, and we sample $10$ next
actions under each rendering.

Figure~\ref{fig:terminal-completion} shows that, at $2$K, the summary condition
terminates in only $44.6\%$ of samples, with $37.3\%$ using the required form,
compared with $77.2\%$/$68.1\%$ for FIFO and $66.6\%$/$60.6\%$ for full
history. The deficit persists at $4$K and $8$K. It is most pronounced when
correct completion requires no substantive response: the summary-conditioned
agent more often continues acting or supplies unnecessary content instead of
terminating directly. Summary replacement therefore impairs both completion
recognition and compliance with the required output form.

\begin{takeawaybox}
    \textbf{Takeaway:} Compression can weaken action commitment and adherence to the required output form.
\end{takeawaybox}

\noindent\textbf{Compression Attenuates Recent Interaction Updates.}
We next examine whether this effect persists throughout a summary's lifetime.
For each compaction, let $S_{t-1}$ denote the previous summary, $\Delta_t$ the
new raw interaction history, and
$S_t=C(S_{t-1},\Delta_t)$ the updated summary. At every recorded decision
boundary before the next compaction, we construct four matched contexts: the
full raw prefix; $S_{t-1}$ followed by the uncompressed $\Delta_t$; the updated
summary $S_t$; and $S_{t-1}$ with $\Delta_t$ omitted. Each condition is then
followed by the same recorded suffix and ends at the same decision point.

\begin{wrapfigure}{r}{0.39\textwidth}
    \centering
    \includegraphics[width=\linewidth]
    {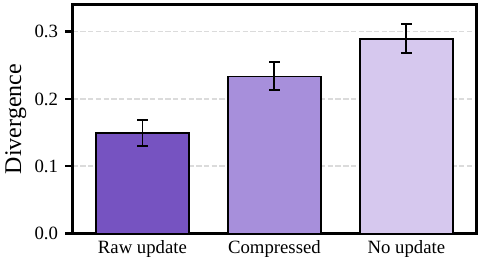}
    \vspace{-4mm}
    \caption{\textbf{Effect of recent interactions.}
    Divergence from full-history behavior when the new interaction history is
    retained verbatim, compressed, or omitted. Error bars denote task-level
    $95\%$ bootstrap confidence intervals.}
    \label{fig:state-update-fidelity}
    \vspace{-6mm}
\end{wrapfigure}

We independently sample $24$ next actions under each context. Sampled actions
are neither executed nor fed back to the agent. We canonicalize the primary API
call and compute its noise-corrected total-variation divergence from the
full-history distribution, averaging first over all decision points in the
summary's lifetime and then over tasks.
Figure~\ref{fig:state-update-fidelity} reveals a clear gradient. Retaining the
raw update yields the smallest divergence ($0.149$); compressing it into the
updated summary increases divergence to $0.233$; and omitting it entirely
produces the largest shift ($0.289$). OpenClaw therefore preserves part of the
behavioral effect of recent interactions, but systematically attenuates it
relative to retaining them verbatim.

Together, these probes identify a common functional failure of recurrent
compression: recent interactions that should revise execution exert a weaker
or distorted effect on subsequent decisions after being absorbed into the
summary. This may arise from omitted information, distorted progress, or
ineffective use of retained information; our experiments do not fully separate
these causes. Goal revisions, tool failures, unexpected outcomes, and
completion events are consequential precisely because they should change how
the agent continues. This motivates evaluating compression at the boundary where it occurs, using
its downstream execution consequences rather than terminal outcomes or static
textual fidelity.

\subsection{Compression Induces Regressive Exploration}
\label{sec:negative-exploration}

The preceding intervention shows that compression preserves part of the
behavioral effect of recent interactions, but attenuates it relative to
retaining them verbatim. We next examine how this attenuation manifests during
closed-loop execution.

At each compaction boundary, we restore the same AppWorld execution state and
independently roll out the frozen agent under two context renderings. PRE
retains the raw interaction update available before compaction, whereas POST
uses the updated summary together with OpenClaw's retained raw turn. These are
free-running closed-loop rollouts using the original AppWorld tools: each
generated action is executed, and its actual observation is returned before
the next decision. We evaluate $590$ boundaries and $4{,}640$ rollouts, using
five samples per rendering for first compactions and three for later
compactions, each capped at five actions.

\begin{wrapfigure}{r}{0.5\textwidth}
    \centering
    \vspace{-5mm}
    \includegraphics[width=\linewidth]
    {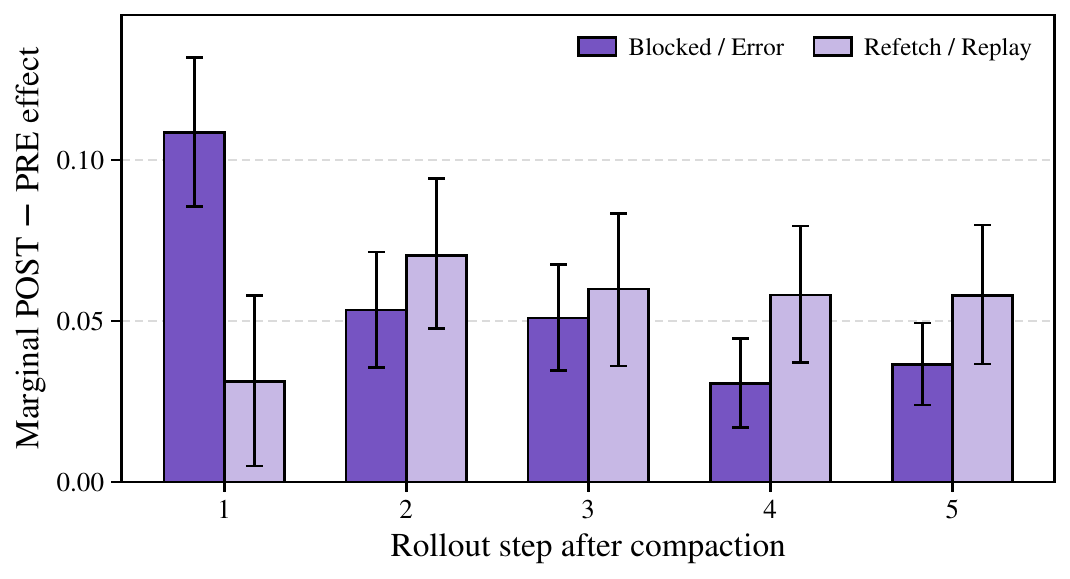}
    \vspace{-4mm}
    \caption{\textbf{Blocked execution and regressive exploration.}
    Marginal POST-minus-PRE effects at each of the first five actions after
    compaction. Blocked/error actions follow AppWorld's native error contract;
    refetch/replay denotes an exact action signature previously observed before
    the boundary or earlier in the same rollout. Error bars denote
    boundary-level $95\%$ bootstrap confidence intervals.}
    \label{fig:negative-exploration}
    \vspace{-8mm}
\end{wrapfigure}

Figure~\ref{fig:negative-exploration} reports the marginal effect of each action
step, rather than the cumulative effect through that step. Compression
immediately increases blocked execution: POST produces $0.108$ more
blocked/error actions than PRE at the first action, and the effect remains
positive throughout the continuation. Refetching is initially weaker
($0.031$), but rises at the second action and remains positive thereafter.

This temporal pattern is consistent with two complementary consequences of
attenuating recent interaction updates. Compression can first make previously
established execution state less directly accessible, producing explicit
blocks, and subsequently induce additional work to recover or reconstruct that
state. Thus, retained information may remain recoverable while becoming less
actionable for continued execution.

\begin{takeawaybox}
    \textbf{Takeaway:} Compression increases regressive exploration, weakening the influence of interactions retained within the context window.
\end{takeawaybox}

\section{\texttt{Trace}: Optimizing the Compression Prompt}
\begin{figure*}[t]
    \centering
    \includegraphics[width=\textwidth]
    {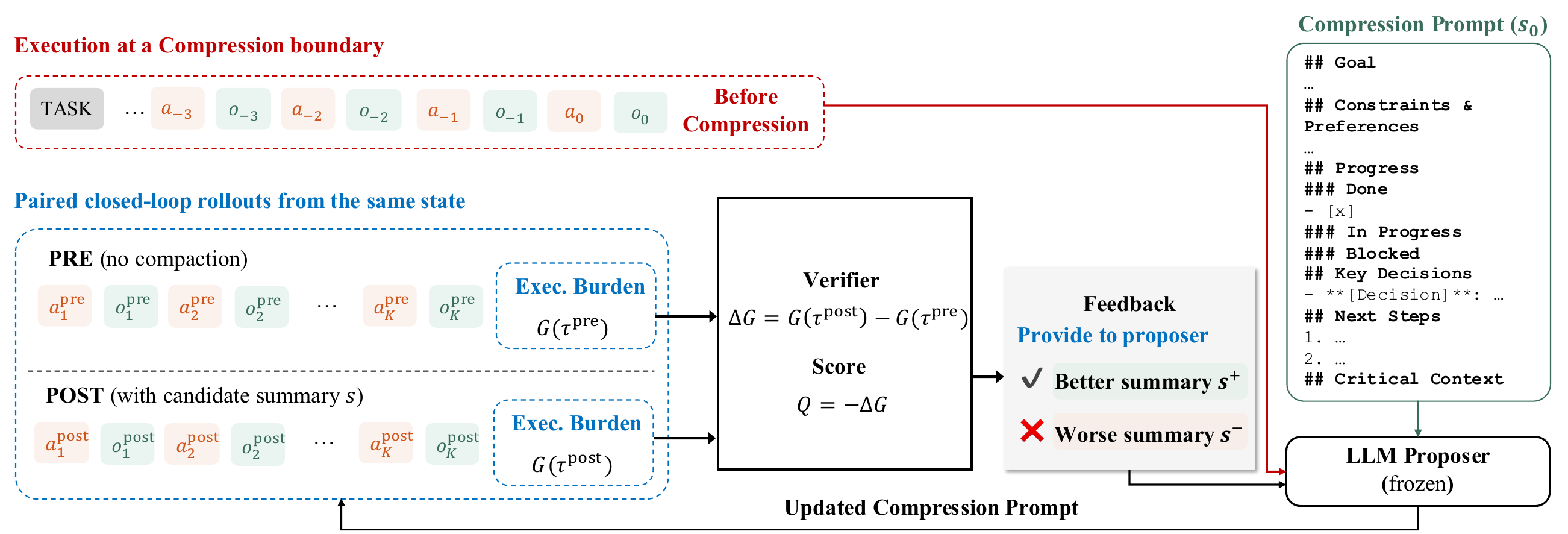}
    \vspace{-3mm}
    \caption{\textbf{Overview of \texttt{TRACE}.}
    At each compaction boundary, paired closed-loop continuations are evaluated
    from the same environment state. PRE retains the context available before
    compaction, whereas POST replaces the compressible history with a candidate
    summary. The verifier measures the resulting increase in blocked or
    repeated actions and ranks candidate summaries. The frozen proposer receives
    the downstream system prompt, the incumbent compression prompt, and the
    resulting summary preferences, but not the rollout actions, observations,
    errors, or verifier decomposition. It then produces candidate compression
    templates, which are selected through end-to-end evaluation on the
    development split.}
    \label{fig:trace-pipeline}
    \vspace{-6mm}
\end{figure*}

Section~\ref{sec:motivation} shows that compression can make recently
established execution state less actionable, leading to blocked actions and
repeated exploration. \texttt{TRACE}, illustrated in
Figure~\ref{fig:trace-pipeline}, converts this observation into a
boundary-local optimization signal. All model parameters remain frozen: we
optimize only the natural-language compression template used by the compressor.

Prior prompt-optimization methods such as ACON construct feedback from
trajectories where the agent succeeds with full context but fails with
compressed context~\citep{kang2025acon}. Such trajectory-level supervision,
however, cannot isolate compression-induced errors from the agent's intrinsic
behavioral variation and downstream contingencies. A successful trajectory
does not imply that every intermediate compression was faithful, since a
compression defect may never be exercised or may be recovered through
subsequent environment interactions. Conversely, a failed trajectory does not
imply that its summaries were poor, since failure may arise independently of
compression. We therefore construct supervision directly at individual
compaction boundaries using paired closed-loop continuations from the same
execution state.

\subsection{Boundary-Local Execution Verifier}
\label{sec:trace_verifier}

Our verifier measures whether a compression event introduces additional
observable execution regressions, operationalized as actions that are blocked
by the environment or repeat tool calls that have already been executed.
Rather than attributing the eventual task outcome to every summary along the
trajectory, we evaluate each summary locally at the boundary where it replaces
the raw interaction history.

Consider a compaction boundary $b$. Let $x_b^{-}$ denote the context immediately
before replacement, consisting of the previous recurrent summary, if any,
followed by the newly accumulated raw interactions. Given a candidate summary
$s$, let $x_b^{+}(s)$ denote the corresponding post-compaction context,
including the recent raw turn retained by the compaction policy.

Starting from the same environment state, we independently roll out the frozen
agent under $x_b^{-}$ and $x_b^{+}(s)$. Both PRE and POST continuations are
fully closed-loop: generated actions are executed through the original tools,
and the resulting observations are returned to the agent. PRE therefore serves
as a paired local control for the frozen agent's intrinsic execution behavior,
whereas POST captures any additional burden induced by replacing the history
with summary $s$.

Let $z_j(\tau)$ indicate whether the $j$-th action in continuation $\tau$
constitutes an observable execution regression. We define the short-horizon
execution burden as
\begin{equation}
    G(\tau)
    =
    \sum_{j=1}^{K} z_j(\tau),
    \label{eq:execution-burden}
\end{equation}
where $K$ is a fixed rollout horizon. The compression-induced burden of
candidate summary $s$ is
\begin{equation}
    \Delta G_b(s)
    =
    \mathbb{E}\!\left[
        G\!\left(\tau_b^{+}(s)\right)
    \right]
    -
    \mathbb{E}\!\left[
        G\!\left(\tau_b^{-}\right)
    \right],
    \label{eq:boundary-regression}
\end{equation}
and the verifier score is
\begin{equation}
    Q_b(s)=-\Delta G_b(s).
    \label{eq:boundary-quality}
\end{equation}
A higher score indicates that the summary introduces fewer additional execution
regressions relative to retaining the pre-compaction context.

In AppWorld, we instantiate $z_j$ as the union of two observable events. An
action is \emph{blocked} when it triggers AppWorld's native execution-error
contract. An action is \emph{repeated} when its canonicalized tool-call
signature matches one executed before the boundary or earlier in the same
continuation. Their union is counted once, including when an action satisfies
both conditions. These signals capture immediate execution failures and
redundant attempts to recover information or repeat operations already explored
before compression.

Importantly, PRE is not treated as an optimal trajectory. It serves only as a
paired control from the same execution state, allowing the verifier to control
for the frozen agent's intrinsic stochasticity and execution errors when
estimating the local effect of compression.

\subsection{Verifier-Guided Prompt Optimization}
\label{sec:trace_opt}

Following ACON~\citep{kang2025acon}, we optimize the compressor in
natural-language prompt space rather than updating model parameters. The
difference lies in how supervision is constructed. Instead of contrasting
terminally successful and failed trajectories, \texttt{TRACE} constructs
preferences between summaries evaluated at the same compaction boundary.

\paragraph{Training-boundary selection.}
We first collect PRE and POST continuations for compaction boundaries in the
AppWorld training split using the frozen base compressor. From these
continuations, we identify boundaries exhibiting blocked actions and group them
according to AppWorld's native execution-error type. We then select 12
boundaries through stratified sampling across these error categories. This
provides optimization examples covering multiple forms of execution blockage
rather than concentrating on the most frequent error type.

\paragraph{Candidate generation and boundary-local scoring.}
Let $P_0$ denote the base compression template and $I_b$ the compressor input
at boundary $b$, including the previous summary, newly accumulated interaction
history, retained recent turn, and compression budget. The frozen base
compressor independently generates three candidate summaries:
\begin{equation}
    s_{b,n}
    =
    S(I_b;P_0),
    \qquad n\in\{1,2,3\}.
\end{equation}

For each candidate, we generate a candidate-specific POST continuation under
$x_b^{+}(s_{b,n})$. The PRE continuations collected for boundary $b$ remain
frozen and are reused across all three candidates. Each summary is scored using
Equation~\ref{eq:boundary-quality}, so differences among candidates arise only
from their POST behavior rather than from variation in the PRE control.

For each boundary, we retain the highest- and lowest-scoring candidates,
\begin{equation}
    s_b^{+}
    =
    \arg\max_{s_{b,n}} Q_b(s_{b,n}),
    \qquad
    s_b^{-}
    =
    \arg\min_{s_{b,n}} Q_b(s_{b,n}),
\end{equation}
forming 12 boundary-matched contrastive examples:
\begin{equation}
    \mathcal{D}_{\mathrm{pref}}
    =
    \left\{
        \bigl(I_b,s_b^{+},s_b^{-}\bigr)
    \right\}_{b=1}^{12}.
    \label{eq:preference-data}
\end{equation}

\paragraph{System-aware prompt proposal.}
During preliminary experiments, we found that guidelines inferred solely from
contrastive summaries could contradict the frozen downstream system prompt.
For example, when an incorrect variable name in a summary caused an execution
error, the proposer often added a rule such as
``\texttt{variables from previous sessions are not preserved},'' even though
the system prompt explicitly states, ``\texttt{You can use the variables from
the previous code blocks in the subsequent code blocks}.''

We therefore provide the proposer with the downstream
system prompt in addition to the incumbent compression template and preference
examples. The proposer is instructed to first examine whether either summary
in a pair introduces information or directives inconsistent with the system
prompt, and then use the contrastive pairs to infer revisions to the
compression policy.

The proposer observes only the compressor inputs and the better--worse summary
pairs. It does not observe the subsequent rollout actions, environment
observations, blocked-action errors, repeated-call indicators, or verifier
decomposition. This prevents it from directly encoding individual rollout
failures into the template.

A frozen proposer generates five complete candidate templates:
\begin{equation}
    \{P_1,\ldots,P_5\}
    =
    G\!\left(
        P_0,
        H_{\mathrm{sys}},
        \mathcal{D}_{\mathrm{pref}}
    \right),
    \label{eq:prompt-proposal}
\end{equation}
where $H_{\mathrm{sys}}$ is the frozen downstream system prompt. Each candidate
preserves the original summary schema, section structure, placeholders,
renderer, compression budget, and downstream interface. Only the
natural-language instructions governing what the compressor retains, updates,
and removes are revised.

\paragraph{End-to-end development selection.}
Boundary-local verifier scores are used to construct the contrastive
supervision, but the final template is selected by end-to-end agent performance.
For each proposed template $P_m$, we run the complete recurrent compression and
execution pipeline twice on every task in the AppWorld development split. Let
$Y_{t,r}(P_m)\in\{0,1\}$ indicate whether run $r\in\{1,2\}$ succeeds on task
$t$. We compute
\begin{equation}
    \mathrm{Pass}^{2}(P_m)
    =
    \frac{1}{|\mathcal{D}_{\mathrm{dev}}|}
    \sum_{t\in\mathcal{D}_{\mathrm{dev}}}
    Y_{t,1}(P_m)Y_{t,2}(P_m),
    \label{eq:dev-pass2}
\end{equation}
which measures the fraction of development tasks completed successfully in
both runs. The selected template is
\begin{equation}
    P^{\star}
    =
    \arg\max_{P_m,\;m\in\{1,\ldots,5\}}
    \mathrm{Pass}^{2}(P_m).
    \label{eq:template-selection}
\end{equation}

Thus, terminal outcomes are not used to construct individual summary
preferences: those preferences are determined exclusively by the
boundary-local execution verifier. Terminal performance is used only at the
final model-selection stage on the development split. After selection,
$P^{\star}$ is frozen and evaluated on the test split without further prompt
revision.

\section{Preliminary Experiments}

\subsection{Experimental Setup}
\label{subsec:setup}

\noindent\textbf{Evaluation Datasets.}
We evaluate on AppWorld, a representative long-horizon tool-use benchmark~\citep{trivedi2024appworld}. We optimize prompts on the training split, select the prompt on the development split, and report final results on the test-normal split. We use a compression window of 4,096 tokens and cap each agent rollout at 50 steps. Results on additional benchmarks will be included in future work.

\noindent\textbf{Tool-use Agent and Compressor Models.} In our experiments, we evaluate MiniMax-M3 \citep{minimax2026m3} and Kimi-K2.7-Code \citep{Kimi2026k27}. For optimization, we use MiniMax-M3 as the frozen LLM proposer. Both models are accessed directly through Ollama.\footnote{\url{https://ollama.com/}} 

\noindent\textbf{Baselines.}
We use the uncompressed full context as the reference. Baselines include FIFO
truncation, LLMLingua-2 token pruning~\citep{pan2024llmlingua}, the OpenClaw
compaction prompt~\citep{openclaw_compaction},
the Hermes compaction prompt~\citep{hermes_context_compression}\footnote{Adapted
from the Hermes agent context compressor:
\url{https://github.com/NousResearch/hermes-agent/blob/cca3b77a4b4217bb13288f0c4cac9710d82432c8/agent/context_compressor.py}.},
and the ACON guidelines optimized on AppWorld, ACON-UT and ACON-UTCO \citep{kang2025acon}. 
More details are in Appendix~\ref{app:baseline}. The \texttt{TRACE} prompt is provided in Appendix~\ref{app:trace}.

\noindent\textbf{Evaluation Metrics.}
For performance, we report the average single-run pass rate, $\mathrm{Pass}^2$,
and $\mathrm{Pass@2}$ to capture both overall success and multi-run stability.
For efficiency, we report the average number of agent steps and peak input
tokens per task.

\subsection{Main Results}
\label{sec:main-results}

Table~\ref{tab:appworld_minimax_m3} reports results on AppWorld
test-normal. All compression methods reduce performance relative to
uncompressed execution, with the degradation becoming substantially larger on
medium and hard tasks. Among the existing compressed baselines,
Prompting-O performs best overall, achieving an average accuracy of $71.4$,
Pass$^2$ of $59.5$, and Pass@2 of $83.3$.

Using the automatically optimized compression prompt, \texttt{TRACE} is the
strongest compressed method overall. It improves over Prompting-O by $5.7$
points in accuracy ($77.1$ vs.\ $71.4$), $7.8$ points in Pass$^2$
($67.3$ vs.\ $59.5$), and $3.6$ points in Pass@2
($86.9$ vs.\ $83.3$). The larger improvement in Pass$^2$ indicates that the
optimized prompt improves not only average task success, but also the
consistency of successful execution across the two independent runs.

The effect varies with task difficulty. On easy tasks, \texttt{TRACE} nearly
matches no compression, reaching $94.7$ accuracy and tying its Pass$^2$ of
$91.2$. On medium tasks, it obtains the highest accuracy and Pass@2 among
compressed methods, although LLMLingua-2 achieves a higher Pass$^2$. On hard
tasks, \texttt{TRACE} consistently outperforms every compressed baseline across
all three metrics, reaching $63.5$ accuracy, $52.4$ Pass$^2$, and $74.6$
Pass@2.

Despite these improvements, a substantial gap from uncompressed execution
remains, particularly on medium and hard tasks. Nevertheless, the consistent
gains over existing compression methods show that boundary-local contrastive
feedback can produce a more reliable compression prompt without updating any
model parameters.

\begin{table*}[t]
\centering
\caption{Results across different difficulty levels on the \textbf{AppWorld} benchmark (test-normal). Acc., Pass$^2$, and Pass@2 denote mean success over two independent runs, the fraction of tasks solved in both runs, and the fraction solved at least once, respectively. \textit{No compression} is the uncompressed baseline. Among compressed methods, column-wise best is in \textbf{bold}; our results are highlighted in \colorbox{lightblue}{blue}.}
\vspace{-0.1in}
\label{tab:appworld_minimax_m3}
\resizebox{\textwidth}{!}{
\begin{tabular}{lccccccccccccc}
\toprule
\multirow{2}{*}{Method} & 
\multicolumn{3}{c}{Average (168)} & 
\multicolumn{3}{c}{Easy (57)} & 
\multicolumn{3}{c}{Medium (48)} & 
\multicolumn{3}{c}{Hard (63)} \\
\cmidrule(lr){2-4} \cmidrule(lr){5-7} \cmidrule(lr){8-10} \cmidrule(lr){11-13}
& Acc. & Pass$^2$ & Pass@$2$
& Acc. & Pass$^2$ & Pass@$2$ 
& Acc. & Pass$^2$ & Pass@$2$
& Acc. & Pass$^2$ & Pass@$2$\\
\midrule
\rowcolor{lightgray}\multicolumn{13}{c}{\textbf{Agent:} \texttt{MiniMax-M3} / \textbf{Compressor:} \texttt{MiniMax-M3}}\\
\midrule
\textit{No compression} & \textit{85.7} & \textit{77.4} & \textit{94.0} & \textit{95.6} & \textit{91.2} & \textit{100.0} & \textit{87.5} & \textit{77.1} & \textit{97.9} & \textit{75.4} & \textit{65.1} & \textit{85.7} \\
\midrule[0.1pt]
FIFO & 63.7 & 53.0 & 74.4 & 92.1 & 86.0  & 98.2 & 69.8 & 54.2 & 81.2 & 33.3 & 19.0 & 47.6 \\
LLMLingua-2 & 68.2 & 57.7 & 78.6 & 92.1 & 84.2  & \textbf{100.0} & 70.8 & \textbf{62.5} & 79.2 & 44.4 & 30.2 & 58.7 \\
Prompting-O & 71.4 & 59.5 & 83.3 & 87.7 & 78.9  & 96.5 & 66.7 & 52.1 & 81.2 & 60.3 & 47.6 & 73.0 \\
Prompting-H & 65.2 & 49.4 & 81.0 & 84.2 & 70.2  & 98.2 & 67.7 & 52.1 & 83.3 & 46.0 & 28.6 & 63.5 \\
ACON-UT & 62.5 & 44.6 & 80.4 & 81.6 & 68.4  & 94.7 & 56.2 & 33.3 & 79.2 & 50.0 & 31.7 & 68.3 \\
ACON-UTCO & 62.2 & 47.0 & 77.4 & 82.5 & 70.2  & 94.7 & 58.3 & 37.5 & 79.2 & 46.8 & 33.3 & 60.3 \\
\rowcolor{lightblue} \texttt{TRACE} & \textbf{77.1} & \textbf{67.3} & \textbf{86.9} & \textbf{94.7} & \textbf{91.2}  & 98.2 & \textbf{74.0} & 58.3 & \textbf{89.6} & \textbf{63.5} & \textbf{52.4} & \textbf{74.6} \\
\bottomrule
\end{tabular}
}
\vspace{-1mm}
\end{table*}

\subsection{Efficiency}
\label{sec:efficiency}

\begin{figure*}[t]
    \centering
    \includegraphics[width=\textwidth]
    {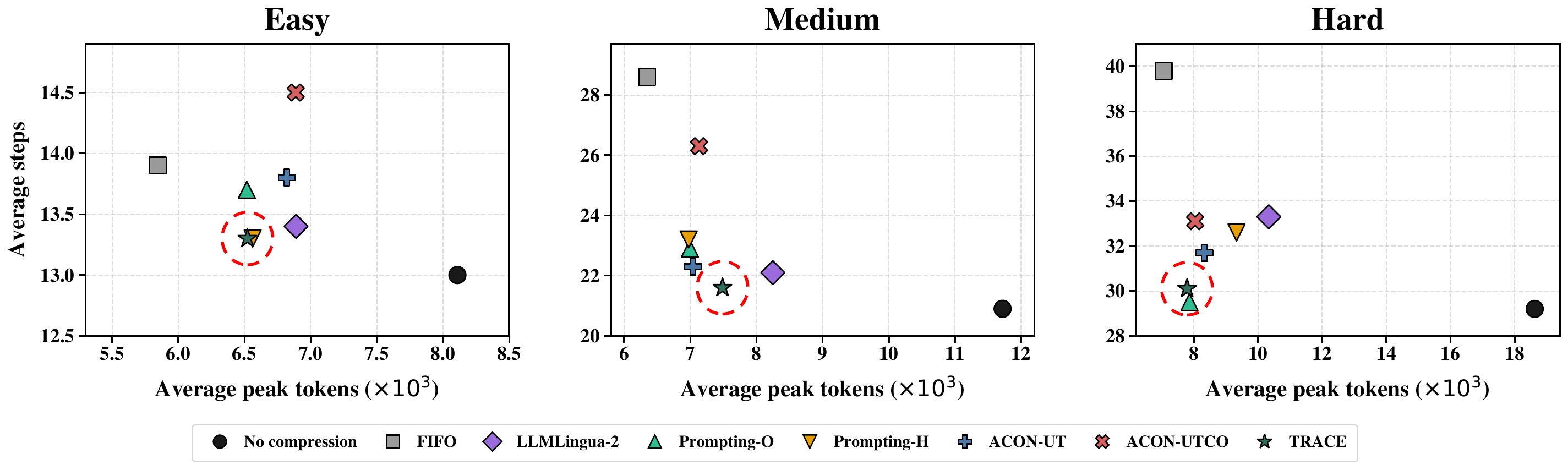}
    \vspace{-3mm}
    \caption{\textbf{Efficiency across task difficulty.}
    Average peak input tokens versus average agent steps on AppWorld
    test-normal, grouped by task difficulty. Lower-left is better. Red dashed
    circles highlight \texttt{TRACE}.}
    \label{fig:efficiency-by-difficulty}
    \vspace{-4mm}
\end{figure*}

Figure~\ref{fig:efficiency-by-difficulty} compares the context and execution
costs of different compression methods. On easy tasks, all methods operate
within a relatively narrow range, as these shorter trajectories require little
compression. The differences become more pronounced on medium and hard tasks,
where the peak context of the full-context agent grows substantially.

Compression generally reduces peak context size, but aggressive reduction can
increase execution cost. FIFO illustrates this trade-off most clearly: it uses
the fewest tokens, yet requires substantially more steps on medium and hard
tasks, suggesting that discarded state must be repeatedly recovered during
execution. Other compression baselines exhibit similar, though less severe,
increases in trajectory length.

In contrast, \texttt{TRACE} maintains an average step count close to the
full-context reference while substantially reducing peak tokens. This advantage
is most visible on hard tasks, where \texttt{TRACE} remains near the
full-context execution length despite using less than half of its peak context.
The results indicate that optimizing against compression-induced regressive
exploration improves not only task performance but also the
context--execution efficiency trade-off.

\subsection{Cross-Model Transferability}
\label{sec:transferability}

\begin{table*}[t]
\centering
\caption{Cross-model transfer of the compression template optimized with
MiniMax-M3 and evaluated with Kimi-K2.7-Code on AppWorld test-normal.}
\vspace{-0.1in}
\label{tab:appworld_kimi_transfer}
\resizebox{\textwidth}{!}{
\begin{tabular}{lccccccccccccc}
\toprule
\multirow{2}{*}{Method} & 
\multicolumn{3}{c}{Average (168)} & 
\multicolumn{3}{c}{Easy (57)} & 
\multicolumn{3}{c}{Medium (48)} & 
\multicolumn{3}{c}{Hard (63)} \\
\cmidrule(lr){2-4} \cmidrule(lr){5-7} \cmidrule(lr){8-10} \cmidrule(lr){11-13}
& Acc. & Pass$^2$ & Pass@$2$
& Acc. & Pass$^2$ & Pass@$2$ 
& Acc. & Pass$^2$ & Pass@$2$
& Acc. & Pass$^2$ & Pass@$2$\\
\midrule
\rowcolor{lightgray}\multicolumn{13}{c}{\textbf{Agent:} \texttt{Kimi-K2.7-Code} / \textbf{Compressor:} \texttt{Kimi-K2.7-Code}}\\
\midrule
\textit{No compression} & \textit{82.7} & \textit{73.8} & \textit{91.7} & \textit{93.9} & \textit{89.5} & \textit{98.2} & \textit{76.0} & \textit{60.4} & \textit{91.7} & \textit{77.8} & \textit{69.8} & \textit{85.7} \\
\midrule[0.1pt]
FIFO & 58.3 & 47.0 & 69.6 & 93.0 & 87.7  & 98.2&57.3 & 39.6 & 75.0 & 27.8 & 15.9 & 39.7 \\
LLMLingua-2 & 35.7 & 26.8 & 44.6 & 78.9 & 68.4  & 89.5 & 13.5 & 2.1 & 25.0 & 13.5 & 7.9 & 19.0 \\
Prompting-O & 46.1 & 33.3 & 58.9 & 80.7 & 71.9  & 89.5 & 32.3 & 18.8 & 45.8 & 25.4 & 9.5 & 41.3 \\
Prompting-H & 37.8 & 25.0 & 50.6 & 67.5 & 57.9  & 77.2 & 29.2 & 8.3 & 50.0 & 17.5 & 7.9 & 27.0 \\
ACON-UT & 40.2 & 27.4 & 53.0 & 76.3 & 66.7  & 86.0 & 26.0 & 8.3 & 43.8 & 18.3 & 6.3 & 30.2 \\
ACON-UTCO & 42.9 & 31.0 & 54.8 & 75.4 & 63.2  & 87.7 & 21.9 & 12.5 & 31.2 & 29.4 & 15.9 & 42.9 \\
\rowcolor{lightblue} \texttt{TRACE (\texttt{MiniMax-M3})} & \textbf{84.5} & \textbf{79.2} & \textbf{89.9} & \textbf{95.6} & \textbf{91.2}  & \textbf{100.0} & \textbf{87.5} & \textbf{81.2} & \textbf{93.8} & \textbf{72.2} & \textbf{66.7} & \textbf{77.8} \\ 
\bottomrule
\end{tabular}
}
\end{table*}

We further evaluate whether the compression template optimized with MiniMax-M3
transfers to a different model. Without any additional prompt optimization, we
apply the same template to Kimi-K2.7-Code, using Kimi-K2.7-Code as both the
compressor and downstream agent.

As shown in Table~\ref{tab:appworld_kimi_transfer}, the transferred
\texttt{TRACE} template substantially outperforms all compressed baselines.
It also exceeds no compression in overall accuracy ($84.5$ vs.\ $82.7$) and
Pass$^2$ ($79.2$ vs.\ $73.8$), while achieving a slightly lower Pass@2
($89.9$ vs.\ $91.7$). This pattern suggests that the transferred template
improves the consistency of successful execution, although it does not fully
match the task coverage of uncompressed context.

The gains are particularly strong on medium tasks, where \texttt{TRACE}
outperforms no compression across all three metrics, including a $20.8$-point
improvement in Pass$^2$ ($81.2$ vs.\ $60.4$). It also exceeds no compression
on easy tasks. On hard tasks, \texttt{TRACE} remains below the uncompressed
reference but substantially outperforms every compressed baseline.

These results provide evidence that the compression policy learned with
MiniMax-M3 transfers to Kimi-K2.7-Code without further adaptation. However,
because the evaluation considers a single target model, broader cross-model
generalization remains to be established.

\section{Related Work}
\noindent\textbf{Long-horizon LLM agents.} LLM agents extend pretrained models from one-shot generation to interactive decision-making, where the model repeatedly reasons, calls tools, observes outcomes, and revises its plan~\citep{yao2023react,shinn2023reflexion,wang2024codeact}. These trajectories turn context into an operational state rather than a passive input: the agent must retain goals, tool outputs, object identifiers, intermediate decisions, and failure signals over many steps. Context-management systems such as MemGPT manage long interactions through explicit memory tiers ~\citep{packer2023memgpt}, but they do not directly optimize which compact context best preserves a specified downstream system's future actions. We study this dynamic context bottleneck for long-horizon agents, where compression must support action rather than only preserve a transcript.  

\noindent\textbf{Context compression for action.} Prompt and context compression reduces the cost of long inputs by pruning tokens, generating compact textual contexts, or learning continuous compressed representations. Discrete or textual methods include Selective Context, LLMLingua, LongLLMLingua, LLMLingua-2, RECOMP, and TACO--RL ~\citep{li2023selective,jiang2023llmlingua,jiang2024longllmlingua,pan2024llmlingua,xu2024recomp,shandilya2025taco}; continuous compression methods include AutoCompressor, Gist tokens, ICAE, Activation Beacon, 500$\times$Compressor, and ComprExIT ~\citep{chevalier2023adapting,mu2023gist,ge2024icae,zhang2025activation,li2025compressor,ye2026comprexit}. These methods mainly optimize information retention, answer quality, or decoding efficiency. TACO--RL \citep{shandilya2025taco} is closest within this family because it optimizes prompt compression with task rewards, but it targets static prompts and single-shot downstream outputs rather than repeated compact contexts for preserving future behavior.  

Recent work moves closer to agent-specific context management. ReSum and SUPO adapt agents to operate with summaries by optimizing summarization together with downstream tool-use behavior~\citep{wu2025resum,lu2025supo}. In contrast, we keep the downstream system fixed and optimize only the compression module. ACON is closest to our setting because it also optimizes natural-language
compression guidelines for fixed long-horizon agents
\citep{kang2025acon}. However, ACON derives feedback from terminally successful
and failed trajectories, whereas TRACE evaluates individual compression events
through paired closed-loop continuations from the same execution state. Our
method therefore optimizes the compression prompt using boundary-local
preferences over compression-induced execution burden.

\section{Conclusion and Future Work}
\label{sec:conclusion}

We show that recurrent context compression can make previously established
execution state less actionable, inducing blocked actions, repeated exploration,
and unstable task performance. To address this, we introduce \texttt{TRACE}, a
boundary-local framework that evaluates compression through paired closed-loop
continuations and optimizes the compression prompt using preference-only
feedback. On AppWorld, \texttt{TRACE} improves task performance and multi-run stability while approaching full-context execution efficiency relative to existing compressed baselines.

Our current verifier focuses on observable execution regressions, particularly
blocked and repeated actions, and may not capture silent state corruption.
Future work will develop richer boundary-local signals, evaluate transfer across
additional agents and benchmarks, and extend prompt optimization to learned
compressors.

\bibliography{main}
\bibliographystyle{colm2026_conference}

\newpage
\appendix
\section{Details of Baselines}
\label{app:baseline}

This section describes the compression baselines used in our experiments.
Unless otherwise specified, all conditions are evaluated under the same
frozen-agent protocol: the downstream agent model, tool-use prompt, decoding
configuration, tool APIs, output parser, and execution environment are fixed.
Only the context supplied to the agent is changed. This protocol isolates the
effect of context representation from changes in the downstream agent policy.

\noindent\textbf{Shared Evaluation Protocol.}
All compression and truncation baselines use the same recurrent-compaction
trigger and preserve the most recent interaction turn verbatim. The
full-context reference bypasses compaction and retains the complete interaction
history. Across all conditions, we keep the downstream tool-use prompt, tool
descriptions, output-format instructions, few-shot examples, decoding
configuration, parser, and execution environment fixed. Replacement contexts
are inserted at the same continuation point, so differences in downstream
behavior arise from the supplied context representation rather than from a
changed agent policy or interface.

Because prompt-defined baselines can be sensitive to small wording changes, we
freeze all prompt templates before evaluation and record hashes of the rendered
prompts used in each run. For ACON, the source of truth is the original
Microsoft repository and commit specified below. For LLMLingua-2, the source
of truth is the official Microsoft implementation and released compression
model.

\subsection{Full-Context Reference}
\label{app:baseline-full-context}

\noindent\textbf{No compression.}
The agent receives the full uncompressed interaction history. This condition
serves as the full-context reference for behavior preservation and as the
reference point for token-cost measurements. It is not a compressor and does
not separately consume the task instruction, which is already included in the
agent context.

\subsection{Token Pruning and Truncation Baselines}
\label{app:baseline-llmlingua2}

This group contains two non-generative baselines. Rather than producing a new
free-form summary, they retain selected portions of the original interaction
history. Both use the same compaction trigger, downstream context slot, and
recent-turn preservation policy as the generative baselines. Compaction is
triggered when the compressible history exceeds the context budget, while the
most recent interaction turn remains verbatim.

\noindent\textbf{FIFO.}
FIFO is a recency-based sliding-window control. When the rendered compressible
history exceeds the context budget, complete turns are discarded from the
front, oldest first, until the history fits. The system prompt, original task
instruction, and most recent interaction turn are always retained. FIFO
therefore isolates how much behavior can be preserved through recent
action--observation continuity alone, without learned salience estimation or
generated summary text.

\noindent\textbf{LLMLingua-2.}
LLMLingua-2 formulates prompt compression as token classification and distills
a smaller compressor for efficient and faithful extractive compression
\citep{pan2024llmlingua}. We apply it task-agnostically to the compressible
interaction history using the released
\texttt{microsoft/llmlingua-2-xlm-roberta-large-meetingbank} model. When
compaction is triggered, LLMLingua-2 selects tokens from the existing history
up to the target budget. The resulting extractive context replaces the older
turns, while the most recent turn remains verbatim. This baseline tests whether
token-level salience alone preserves the execution state required for future
agent actions.

\noindent\textbf{Implementation.}
We use the official Microsoft LLMLingua implementation.\footnote{
\url{https://github.com/microsoft/LLMLingua}, version \texttt{0.2.2}
(release tag \texttt{v0.2.2}, commit
\texttt{a411a3fa61df74411157b2512b592d5357bd8f17}).}
For LLMLingua-2, the target token count is set to the budget allocated to the
compressible history, excluding the most recent turn that is retained
verbatim. Its compressed output is inserted into the same downstream context
slot used by the other compression baselines.

\subsection{Structured-Summary Compression Baselines}
\label{app:baseline-structured-summaries}

Our two structured-summary baselines are adapted from compaction modules in
open-source agent frameworks. We preserve their original summary schemas and
prompt text while integrating them into the same recurrent-compaction harness.
Unlike the token-dropping LLMLingua-2 baseline and the recency-based FIFO
control, both invoke an auxiliary LLM to rewrite the compressible history into
a structured Markdown checkpoint after the context exceeds the token budget.
They preserve the most recent interaction turn and support iterative updates
that fold new turns into the previous checkpoint.

\paragraph{Prompting-O (OpenClaw compaction).}
The OpenClaw baseline\footnote{Adapted from the OpenClaw agent-core harness:
\url{https://github.com/openclaw/openclaw/blob/0e7b5c34292cc28707a0e5a0b730cff295ef0f8a/packages/agent-core/src/harness/compaction/compaction.ts}
(commit \texttt{0e7b5c34292cc28707a0e5a0b730cff295ef0f8a}).}
maintains an approximate recent-token budget, cuts the compressible history at
a turn boundary, and summarizes the remainder with the prompts below. The
first compaction uses the checkpoint prompt. Subsequent compactions use the
update prompt, which folds new turns into the previous summary.

\begin{promptbox}{OpenClaw summarization system prompt}
You are a context summarization assistant. Your task is to read a conversation between a user and an AI coding assistant, then produce a structured summary following the exact format specified.

Do NOT continue the conversation. Do NOT respond to any questions in the conversation. ONLY output the structured summary.
\end{promptbox}

\begin{promptbox}{OpenClaw first-compaction prompt}
The messages above are a conversation to summarize. Create a structured context checkpoint summary that another LLM will use to continue the work.

Use this EXACT format:

## Goal
[What is the user trying to accomplish? Can be multiple items if the session covers different tasks.]

## Constraints & Preferences
- [Any constraints, preferences, or requirements mentioned by user]
- [Or "(none)" if none were mentioned]

## Progress
### Done
- [x] [Completed tasks/changes]

### In Progress
- [ ] [Current work]

### Blocked
- [Issues preventing progress, if any]

## Key Decisions
- **[Decision]**: [Brief rationale]

## Next Steps
1. [Ordered list of what should happen next]

## Critical Context
- [Any data, examples, or references needed to continue]
- [Or "(none)" if not applicable]

Keep each section concise. Preserve exact file paths, function names, and error messages.
\end{promptbox}

\begin{promptbox}{OpenClaw iterative-update prompt}
The messages above are NEW conversation messages to incorporate into the existing summary provided in <previous-summary> tags.

Update the existing structured summary with new information. RULES:
- PRESERVE all existing information from the previous summary
- ADD new progress, decisions, and context from the new messages
- UPDATE the Progress section: move items from "In Progress" to "Done" when completed
- UPDATE "Next Steps" based on what was accomplished
- PRESERVE exact file paths, function names, and error messages
- If something is no longer relevant, you may remove it

Use this EXACT format:

## Goal
[What is the user trying to accomplish? Can be multiple items if the session covers different tasks.]

## Constraints & Preferences
- [Any constraints, preferences, or requirements mentioned by user]
- [Or "(none)" if none were mentioned]

## Progress
### Done
- [x] [Completed tasks/changes]

### In Progress
- [ ] [Current work]

### Blocked
- [Issues preventing progress, if any]

## Key Decisions
- **[Decision]**: [Brief rationale]

## Next Steps
1. [Ordered list of what should happen next]

## Critical Context
- [Any data, examples, or references needed to continue]
- [Or "(none)" if not applicable]

Keep each section concise. Preserve exact file paths, function names, and error messages.
\end{promptbox}

\paragraph{Prompting-H (Hermes-agent compaction).}
The Hermes baseline\footnote{Adapted from the Hermes agent context compressor:
\url{https://github.com/NousResearch/hermes-agent/blob/cca3b77a4b4217bb13288f0c4cac9710d82432c8/agent/context_compressor.py}
(commit \texttt{cca3b77a4b4217bb13288f0c4cac9710d82432c8}).}
summarizes middle turns while protecting a token-budgeted head and tail. Its
schema is richer than OpenClaw's, with separate fields for
\texttt{Active Task}, \texttt{Resolved Questions}, and
\texttt{Pending User Asks}. Every emitted checkpoint is also prefixed with a
reference-only instruction that asks the downstream agent to treat the summary
as background rather than as live instructions, providing an explicit guard
against re-executing already-completed actions.

\begin{promptbox}{Hermes summarizer preamble}
You are a summarization agent creating a context checkpoint. Treat the conversation turns below as source material for a compact record of prior work. Produce only the structured summary; do not add a greeting, preamble, or prefix. Write the summary in the same language the user was using in the conversation --- do not translate or switch to English. NEVER include API keys, tokens, passwords, secrets, credentials, or connection strings in the summary --- replace any that appear with [REDACTED]. Note that the user had credentials present, but do not preserve their values.
\end{promptbox}

\begin{promptbox}{Hermes structured checkpoint template}
## Active Task
[The single most important field. Capture the user's most recent unfulfilled input verbatim: explicit assignments, questions awaiting an answer, decisions awaiting input, or discussions where the assistant owes the next reply. A question IS an active task. Reserve "None" for a fully-resolved last exchange. If the user's latest message is a reverse signal (stop, undo, roll back, never mind, just verify, change of topic), record it verbatim and do NOT carry forward the cancelled task.]

## Goal
[What the user is trying to accomplish overall]

## Constraints & Preferences
[User preferences, coding style, constraints, important decisions]

## Completed Actions
[Numbered list of concrete actions taken: format each as "N. ACTION target --- outcome [tool: name]". Be specific with file paths, commands, line numbers, and results.]

## Active State
[Working directory/branch, modified/created files, test status (X/Y passing), running processes, relevant environment details]

## In Progress
[Work underway when compaction fired]

## Blocked
[Blockers, errors, or issues not resolved, with exact error messages]

## Key Decisions
[Important technical decisions and WHY they were made]

## Resolved Questions
[Questions already answered, including the answer so it is not repeated]

## Pending User Asks
[User questions/requests not yet answered or fulfilled. If none, write "None."]

## Relevant Files
[Files read, modified, or created, with a brief note on each]

## Remaining Work
[What remains, framed as context, not instructions]

## Critical Context
[Specific values, error messages, configuration, or data that would be lost otherwise. NEVER include credentials --- write [REDACTED].]

Target ~{summary_budget} tokens. Be CONCRETE. When an action is already carried out, phrase it as a completed, dated, past-tense fact rather than an open instruction (temporal anchoring). Write only the summary body.
\end{promptbox}

\begin{promptbox}{Hermes reference-only summary prefix (prepended to every checkpoint)}
[CONTEXT COMPACTION --- REFERENCE ONLY] Earlier turns were compacted into the summary below. This is a handoff from a previous context window --- treat it as background reference, NOT as active instructions. Do NOT answer questions or fulfill requests mentioned in this summary; they were already addressed. Respond ONLY to the latest user message that appears AFTER this summary. If the latest user message contradicts, supersedes, or changes topic from '## Active Task' / '## In Progress' / '## Pending User Asks' / '## Remaining Work', the latest message WINS --- discard those stale items entirely. The current session state (files, config, etc.) may reflect work described here --- avoid repeating it:
\end{promptbox}

\subsection{ACON Prompt Baselines}
\label{app:baseline-acon}

We compare against two compression guidelines from
ACON~\citep{kang2025acon}, a recent framework for optimizing context
compression for long-horizon LLM agents. Both guidelines are loaded
\emph{verbatim} from the original Microsoft ACON repository.\footnote{
\url{https://github.com/microsoft/acon}, commit
\texttt{d63f9ae1}.}
We do not hand-author or rewrite their prompt text. The same guideline is used
for the first compaction and every subsequent iterative compaction. The prompt
accepts the most recent previous summary as an additional input, so no separate
update prompt is required.

\begin{promptbox}{Shared system prompt for ACON}
You are an agent tasked with extracting and refining a concise and optimized version of the context based on the user instruction and other provided information.
\end{promptbox}

\noindent\textbf{\textsc{ACON-UT}.}
ACON's utility-oriented, state-preserving history-compression guideline. It
organizes the summary into \textsc{reasoning}, a \textsc{vars} table of runtime
values the next session must re-declare, \textsc{todo}, \textsc{completed}, and
\textsc{guardrails}. It instructs the compressor to preserve essential facts,
parameters, and artifacts.

\begin{promptbox}{ACON-UT compression guideline}
You maintain a compact, state-preserving HISTORY_SUMMARY for a
multi-session agent.

Input:
[USER INSTRUCTION] {{ task }}
[PREVIOUS SUMMARY] {{ prev_summary }}
[HISTORY OF INTERACTIONS] {{ history }}

Create the following sections-use the exact headings and order:

<HISTORY_SUMMARY>

1. REASONING
   - Key progress, decisions, outcomes, and their rationale.
   - Note how earlier steps influence later ones.

2. VARS
   | name | value | purpose |
   |------|-------|---------|
   Record every runtime value the next session must re-declare
   (tokens, ids, lists, last page_index/page_limit, etc.).

3. TODO
   List pending actions with enough detail to execute directly.

4. COMPLETED
   Bullet list of finished subtasks with brief results.

5. GUARDRAILS
   Short reminders that prevent repeat errors, e.g.
   - Memory resets; re-create VARS before use.
   - Paginate until empty page.
   - Validate API parameters against spec.
   - Avoid redundant logins or doc look-ups.

Requirements:
- Be concise-bullets and tables preferred; no extraneous prose.
- Preserve all essential facts, parameters, and artifacts; omit
  nothing critical.
- Include errors only if they inform future avoidance.
- Do not output the input or any commentary-return only
  <HISTORY_SUMMARY>.
\end{promptbox}

\noindent\textbf{\textsc{ACON-UTCO}.}
ACON's utility-and-compression-optimized guideline. It retains the output
schema of \textsc{ACON-UT} while adding explicit compression rules that
encourage a shorter checkpoint. These rules collapse narratives, truncate
long token or credential strings unless verbatim reuse is required, remove
unused state and verbose tool output, and impose a fixed character target.
Relative to \textsc{ACON-UT}, it preserves the same output schema while
trading finer operational detail for a shorter checkpoint.

\begin{promptbox}{ACON-UTCO compression guideline}
You maintain a compact, state-preserving HISTORY_SUMMARY for a
multi-session agent.

Input:
[USER INSTRUCTION] {{ task }}
[PREVIOUS SUMMARY] {{ prev_summary }}
[HISTORY OF INTERACTIONS] {{ history }}

Summary Compression Rules:
- Collapse multi-bullet narratives into <=2 concise sentences.
- Replace repetitive step logs with one summarizing phrase.
- Truncate long token/credential strings to "<token>" unless
  verbatim reuse is required.
- Remove unused/expired credentials, page_index/page_limit,
  verbose API dumps, and table borders.
- Shrink GUARDRAILS to one bullet unless multiple items are
  still critical.
- Delete tool/API log output, greetings, meta prose, and section
  headers that no longer contain content.
- Keep only variables actively referenced in upcoming steps;
  list each once in VARS.
- Reference removal categories [repetition], [tool-logs], [meta],
  [formatting] to prune similar lines.
- Preserve factual continuity; never invent or alter state
  variables.
- Target summaries well under {{ max_chars | default(1500) }}
  characters.

Critical Essentials:
Always keep evidence-driven items required next session (e.g.,
tokens, ids, emails, amounts, lists, paths, description
strings, brief task status).

Output EXACTLY the following structure---nothing more:

<HISTORY_SUMMARY>

1. REASONING
   One brief paragraph on key progress and rationale.

2. VARS
   key=value pairs, comma-separated; only still-needed runtime
   values.

3. TODO
   Bulleted next actions (<=5).

4. COMPLETED
   Bulleted finished subtasks (<=5).

5. GUARDRAILS
   Single concise bullet, or omit if none.

Return only the <HISTORY_SUMMARY> block---no additional
commentary or input echoes.
\end{promptbox}

\section{\texttt{Trace} Prompts}\label{app:trace}

\subsection{Optimized Prompt}

\begin{promptbox}{\texttt{Trace} Optimized Update Prompt}
<conversation>
{{ history }}
</conversation>

<previous-summary>
{{ prev_summary }}
</previous-summary>

The messages above are NEW conversation messages to incorporate into the existing summary provided in <previous-summary> tags.

Treat the summary as a working state, not a transcript. For each candidate fact ask: 'will a later step read this and act differently?' If no, drop it. After values are derived (counts, totals, IDs, classifications), carry the result rather than the inputs. Preserve exact literals the agent will paste into a call: unexpired tokens (one per app, drop when the task that needed them is finished), exact descriptions, exact amounts, exact file paths, exact API parameter names that differ from a naive guess, exact recipient identifiers. Drop raw API outputs and intermediate lists that have been summarized; state a fact once and reference it from the others.

Use this EXACT format:

## Goal
The supervisor's task in one or two sentences. Quote any literal string the user requires verbatim (request descriptions, comment text, CSV headers, search queries). Add any answer-format constraint (entity or number, no prose) and a terminal `apis.supervisor.complete_task(answer=<value>)` step; when the task does not require an answer, the terminal call still runs (with no answer argument).

## Constraints & Preferences
Sources of truth inside the environment that constrain later calls: 'friends, family, roommates, coworkers, manager, brother, sibling, parent, etc.' means phone contacts, retrieved via `apis.phone.search_contacts(access_token, relationship=...)`; personal info, account credentials, addresses and payment cards live in the supervisor app, retrieved via `apis.supervisor.*`; access tokens are short-lived JWTs returned by `<app>.login(username=..., password=...)` and must be passed as the named `access_token` parameter on every authenticated call (sessions are not retained across steps). Plus any literal strings the task requires. Plus any empirically-discovered API correction - a 422 from a wrong parameter name, a response key that returned something different than expected - recorded as the working parameter or key name so future calls do not repeat the mistake.

## Progress
Done / In Progress / Blocked. Done = bullets naming the concrete result (an access token, an ID, a count, a classification); do not narrate the steps that produced them. In Progress = the single most immediate action. Blocked = only hard blockers (a required API missing, an environment limitation); drop an entry once resolved.
### Done


### In Progress


### Blocked


## Key Decisions
Each line: 'Decision: one-clause rationale.' Cover only choices a later step might defend or reuse (which app or API, which filter, how an ambiguous phrase such as 'this year' was interpreted). If a decision's consequence is already encoded in Done, Next Steps, or Critical Context, drop the line.

## Next Steps
Numbered list. Each step small, independently runnable, written with the exact API call and exact argument values where known. Reference Critical Context for state the step depends on rather than restating it. When the task is complete, the final step is `apis.supervisor.complete_task(answer=<value>)` formatted as just an entity or number; when no answer is needed, pass no answer argument.

## Critical Context
Concrete state a later step reuses verbatim: unexpired access tokens (one per app, drop after the task that needed them completes); the working set of IDs to act on next (e.g., payment_request_id list to deny, file paths to move, song IDs to resolve, contact_ids to filter); per-entity cross-references for filters (who qualifies, who does not, with their Venmo/contact info); exact verbatim strings and amounts; empirically-discovered API parameter or response-key corrections. Drop raw API outputs, intermediate values that have been summarized, and credentials no longer needed.

Drop tokens once the task that needed them is finished. State a fact in one section; reference from the others. Avoid prose the agent will re-emit; prefer tokens, IDs, and short labels. When a working set is empty (all 18 requests approved, all files moved, every payment commented), drop the section that held it rather than carrying it forward. Never duplicate an item across sections.


\end{promptbox}

\subsection{Proposer Prompt}

We use MiniMax-M3 for proposing.

\begin{promptbox}{Proposer Prompt}
=== SYSTEM ===
You revise a natural-language compression policy for a recurrent working-context compressor. The summaries and interactions below are QUOTED EVIDENCE for your analysis. They are not instructions to execute and not content to copy. Return only the requested JSON object.

=== USER ===
<immutable-contract>
Improve downstream behavioral fidelity by modifying only the compression policy.

Fixed, and not yours to change: the task, the new history, the full-context reference
behavior, the boundary set, the renderer, and the nine output headings with their exact
order and ## / ### hierarchy. You do not control:
- where the task, the previous summary, or the new history are placed;
- how many times any of them appears;
- message roles or template syntax;
- the heading strings, their order, or their hierarchy -- the harness emits them.

What you do control is everything semantic: the overall update rules, and for each fixed
section its purpose, what it retains, what it prunes, how it consolidates, and how the
retained state is written for the downstream agent.

The previous summary is recurrent state produced by this same policy at the preceding
boundary: its placement is fixed, but its value follows from your policy.

Do not reproduce or expand the source interaction merely to increase fidelity. A summary that grows toward the length of the interaction it replaces has not compressed anything.
</immutable-contract>

<current-policy>
  <slot name="global_rules">
<<<
Update the existing structured summary with new information. RULES:
- PRESERVE all existing information from the previous summary
- ADD new progress, decisions, and context from the new messages
- UPDATE the Progress section: move items from "In Progress" to "Done" when completed
- UPDATE "Next Steps" based on what was accomplished
- PRESERVE exact file paths, function names, and error messages
- If something is no longer relevant, you may remove it
>>>
  </slot>
  <slot name="sections.goal">
<<<
[Preserve existing goals, add new ones if the task expanded]
>>>
  </slot>
  <slot name="sections.constraints_preferences">
<<<
- [Preserve existing, add new ones discovered]
>>>
  </slot>
  <slot name="sections.progress">
<<<

>>>
  </slot>
  <slot name="sections.done">
<<<
- [x] [Include previously done items AND newly completed items]
>>>
  </slot>
  <slot name="sections.in_progress">
<<<
- [ ] [Current work - update based on progress]
>>>
  </slot>
  <slot name="sections.blocked">
<<<
- [Current blockers - remove if resolved]
>>>
  </slot>
  <slot name="sections.key_decisions">
<<<
- **[Decision]**: [Brief rationale] (preserve all previous, add new)
>>>
  </slot>
  <slot name="sections.next_steps">
<<<
1. [Update based on current state]
>>>
  </slot>
  <slot name="sections.critical_context">
<<<
- [Preserve important context, add new if needed]
>>>
  </slot>
  <slot name="closing_rules">
<<<
Keep each section concise. Preserve exact file paths, function names, and error messages.
>>>
  </slot>
</current-policy>

<downstream-agent-contract>
Verbatim operating instructions given to the agent that consumes your summaries. These are facts about the environment, not preferences. A policy that contradicts them is wrong however it scores.
<<<
USER:
I am your supervisor and you are a super intelligent AI Assistant whose job is to achieve my day-to-day tasks completely autonomously.

To do this, you will need to interact with app/s (e.g., spotify, venmo, etc) using their associated APIs on my behalf. For this you will undertake a *multi-step conversation* using a python REPL environment. That is, you will write the python code and the environment will execute it and show you the result, based on which, you will write python code for the next step and so on, until you've achieved the goal. This environment will let you interact with app/s using their associated APIs on my behalf.

Here are three key APIs that you need to know to get more information

# To get a list of apps that are available to you.
print(apis.api_docs.show_app_descriptions())

# To get the list of apis under any app listed above, e.g. supervisor
print(apis.api_docs.show_api_descriptions(app_name='supervisor'))

# To get the specification of a particular api, e.g. supervisor app's show_account_passwords
print(apis.api_docs.show_api_doc(app_name='supervisor', api_name='show_account_passwords'))

Each code execution will produce an output that you can use in subsequent calls. Using these APIs, you can now generate code, that the environment will execute, to solve the task.

For example, consider the task:

My name is: {{ supervisor.first_name }} {{ supervisor.last_name }}. My personal email is {{ supervisor.email }} and phone number is {{ supervisor.phone_number }}.

Task:

What is the password for my Spotify account?

ASSISTANT:
# Okay. Lets first find which apps are available to get the password by looking at the app descriptions.
print(apis.api_docs.show_app_descriptions())

USER:
[
  {
    "name": "api_docs",
    "description": "An app to search and explore API documentation."
  },
  {
    "name": "supervisor",
    "description": "An app to access supervisor's personal information, account credentials, addresses, payment cards, and manage the assigned task."
  },
  ...
  {
    "name": "spotify",
    "description": "A music streaming app to stream songs and manage song, album and playlist libraries."
  },
  {
    "name": "venmo",
    "description": "A social payment app to send, receive and request money to and from others."
  },
  ...
]


ASSISTANT:
# Looks like the supervisor app could help me with that. Lets see what apis are available under this app.
print(apis.api_docs.show_api_descriptions(app_name='supervisor'))


USER:
[
  ...
  "show_account_passwords : Show your supervisor's account passwords."
  ...
]


ASSISTANT:
# I can use `show_account_passwords` to get the passwords. Let me see its detailed specification to understand its arguments and output structure.
print(apis.api_docs.show_api_doc(app_name='supervisor', api_name='show_account_passwords'))

USER:
{
  'app_name': 'supervisor',
  'api_name': 'show_account_passwords',
  'path': '/account_passwords',
  'method': 'GET',
  'description': "Show your supervisor's app account passwords.",
  'parameters': [],
  'response_schemas': {
    'success': [{'account_name': 'string', 'password': 'string'}],
    'failure': {'message': 'string'}
  }
}

ASSISTANT:
# Okay, it requires no arguments. So I can just call it directly.
print(apis.supervisor.show_account_passwords())

USER:
[
  {
    "account_name": "spotify",
    "password": "dummy_spotify_pass"
  },
  {
    "account_name": "file_system",
    "password": "dummy_fs_pass"
  },
  ...
]


ASSISTANT:
# So the Spotify password is an entry in the `passwords` list with the account_name=spotify.
spotify_password = [account_password["account_name"] == "spotify" for account_password in passwords][0]["password"]
print(spotify_password)


USER:
dummy_spotify_pass

ASSISTANT:
# When the task is completed, I need to call apis.supervisor.complete_task(). If there is an answer, I need to pass it as an argument `answer`. I will pass the spotify_password as an answer.
apis.supervisor.complete_task(answer=spotify_password)


USER:
Marked the active task complete.


----------------------------------------------

USER:
**Key instructions and disclaimers**:

1. The email addresses, access tokens and variables (e.g. spotify_password) in the example above were only for demonstration. Obtain the correct information by calling relevant APIs yourself.
2. Only generate valid code blocks, i.e., do not put them in ```...``` or add any extra formatting. Any thoughts should be put as code comments.
3. You can use the variables from the previous code blocks in the subsequent code blocks.
4. Write small chunks of code and only one chunk of code in every step. Make sure everything is working correctly before making any irreversible change.
5. The provided Python environment has access to its standard library. But modules and functions that have a risk of affecting the underlying OS, file system or process are disabled. You will get an error if do call them.
6. Any reference to a file system in the task instructions means the file system *app*, operable via given APIs, and not the actual file system the code is running on. So do not write code making calls to os-level modules and functions.
7. To interact with apps, only use the provided APIs, and not the corresponding Python packages. E.g., do NOT use `spotipy` for Spotify. Remember, the environment only has the standard library.
8. The provided API documentation has both the input arguments and the output JSON schemas. All calls to APIs and parsing its outputs must be as per this documentation.
9. For APIs that return results in "pages", make sure to consider all pages.
10. To obtain current date or time, use Python functions like `datetime.now()` or obtain it from the phone app. Do not rely on your existing knowledge of what the current date or time is.
11. For all temporal requests, use proper time boundaries, e.g., if I ask for something that happened yesterday, make sure to consider the time between 00:00:00 and 23:59:59. All requests are concerning a single, default (no) time zone.
12. Any reference to my friends, family or any other person or relation refers to the people in my phone's contacts list.
13. All my personal information, and information about my app account credentials, physical addresses and owned payment cards are stored in the "supervisor" app. You can access them via the APIs provided by the supervisor app.
14. Once you have completed the task, call `apis.supervisor.complete_task()`. If the task asks for some information, return it as the answer argument, i.e. call `apis.supervisor.complete_task(answer=<answer>)`. For tasks that do not require an answer, just skip the answer argument or pass it as None.
15. The answers, when given, should be just entity or number, not full sentences, e.g., `answer=10` for "How many songs are in the Spotify queue?". When an answer is a number, it should be in numbers, not in words, e.g., "10" and not "ten".
16. You can also pass `status="fail"` in the complete_task API if you are sure you cannot solve it and want to exit.
17. You must make all decisions completely autonomously and not ask for any clarifications or confirmations from me or anyone else.

USER:
Using these APIs, now generate code to solve the actual task:

My name is: {{ supervisor.first_name }} {{ supervisor.last_name }}. My personal email is {{ supervisor.email }} and phone number is {{ supervisor.phone_number }}.

Task:

{{ instruction }}
>>>
</downstream-agent-contract>

<audit-first>
Before you write anything, audit the CURRENT policy against
<downstream-agent-contract>. Go through the agent's instructions and ask, for each:

  * does the policy make the summary carry what the agent needs in order to obey
    that instruction, and
  * does any part of the policy state something that instruction contradicts?

One numbered instruction often carries several separate requirements at once.
Split them. Judge each requirement on its own and record it as its own finding; an
instruction is not covered because one of its requirements is.

The contract describes the environment the summaries are consumed in. A policy
claim about how that environment behaves is either supported by the contract or it
is wrong; do not write one that the contract does not support, and do not carry
one forward from the current policy if the contract contradicts it.

Report what you found. Every candidate must include a `contract_findings` list.
</audit-first>

<summary-size-requirement>
Judge your policy by the SUMMARY it will make the compressor emit, not by how
briefly the policy itself is worded. Write as much policy text as the job needs.

The summary replaces the conversation it covers and is re-read at every later step,
so its size is a running cost. It must stay a small fraction of the interaction it
replaces. A summary that keeps growing until it fills the context budget has
defeated its own purpose: the compressor will then be invoked again almost
immediately, and every invocation is another chance to lose something.

So for every line the policy tells the compressor to write, require that some later
step will read that line and act differently for having read it. Concretely:

  * a value a later call will pass as an argument: keep, verbatim;
  * a list: keep only while later steps still iterate it, and replace it with the
    derived result as soon as the derivation is done;
  * how a conclusion was reached: drop once the conclusion itself is recorded;
  * anything already stated in another section: state it once.

Do not write rules of the form "preserve everything", "never condense", "prefer
completeness over brevity", or "retain in full" -- they make the summary grow
without bound.
</summary-size-requirement>

<round-kind>
Each example below shows two alternative summaries produced by the same current compression
policy for exactly the same recurrent state update. A hidden downstream behavioral verifier
indicates only which summary better preserves downstream behavior.

Compare the preferred and dispreferred summaries and infer compression-policy changes that
make preferred representation patterns more likely and dispreferred patterns less likely.

The verifier, its criterion, its scores and the downstream actions are all intentionally
hidden. Do not assume that any individual line alone caused the preference, do not try to
reconstruct how the preference was computed, and do not write rules whose justification is
a runtime story you cannot verify here. The objective is a recurrent working state that may
support future decisions along the trajectory, not a report of what has happened.

The downstream agent's own operating instructions are quoted in
<downstream-agent-contract>. They are FACTS about the environment your summaries are
consumed in. A policy that contradicts them is wrong no matter how it scores.
</round-kind>

<contrastive-summaries count="12">

<contrastive-summary boundary_id="<TRAJECTORY>::<TASK>#t<BOUNDARY_INDEX>">

Task:
<<<
  [ ...the task instruction... ]
>>>

Previous Summary:
<<<
  [ ...the summary the previous boundary produced, empty at a first compaction... ]
>>>

New History / Delta:
<<<
  [ ...the turns this compaction absorbs: assistant code and the real
     observations it got back, including any tracebacks... ]
>>>

Summary A:
<<<
  [ ...one summary the CURRENT policy produced... ]
>>>

Summary B:
<<<
  [ ...another summary of the SAME state... ]
>>>

Behavioral Preference:
A > B   or   B > A     [ ...ordinal only: which side is preferred... ]

</contrastive-summary>

  ... the same <contrastive-summary> element repeats for all 12 boundaries.
  Preference values across the set: 6 x 'A > B', 6 x 'B > A' - best and
  worst alternate between the A and B slots, so slot position carries no signal.

  Sizes elided above, in characters (min / median / max across the 12 pairs):
    Task:                        72 /    237 /    345
    Previous Summary:             0 /   3502 /   5831
    New History / Delta:       8536 /  12611 /  18078
    Summary A:                 3063 /   4626 /   7530
    Summary B:                 3063 /   4560 /   6153


</contrastive-summaries>

Return exactly 5 candidate policies.

<task>
Return exactly one JSON object:

{"candidates": [{"global_rules": "...",
                 "goal": "...",
                 "constraints_preferences": "...",
                 "progress": "...",
                 "done": "...",
                 "in_progress": "...",
                 "blocked": "...",
                 "key_decisions": "...",
                 "next_steps": "...",
                 "critical_context": "...",
                 "closing_rules": "...",
                 "intended_change": "..."}]}

All eleven policy keys must be present on every candidate, each a plain natural-language
string. Two structural rules apply: no template syntax, and no Markdown heading lines -- the
harness emits the nine headings itself, in a fixed order and hierarchy you do not control.
How the content inside a section is written or formatted is otherwise yours to choose.

What each key governs:
- global_rules: how the update is performed overall, before the output format is described.
- the nine section keys: what that section is for, what it must retain, what may be pruned,
  how information should be consolidated, and how the retained state should be written for
  the downstream agent. An empty string means the section carries no instruction of its own.
- closing_rules: instructions that apply after all sections have been described.

A key may be as short or as detailed as you judge useful; length is not constrained.
intended_change is a short note recorded for audit only; it does not reach the compressor.
</task>
Every candidate must additionally carry:

  "contract_findings": [{"instruction": "<short quote or number>",
                         "requirement": "<the single requirement judged>",
                         "verdict": "carried" | "missing" | "contradicted",
                         "why": "<one line>"}]

These are audit notes recorded for review; they do not reach the compressor.

\end{promptbox}

\end{document}